\newcommand{\noind}{\hspace*{-1 \parindent}}
\newcommand{\qit}[1]{{\sl ``#1''}}         
\newcommand{\opr}[1]{\mbox{{\em #1}\/}}    
\newcommand{\mtxt}[1]{\mbox{{\em #1}\/}}   
\newcommand{\nlidb}{{\sc Nlidb}}
\newcommand{\nlidbs}{{\sc Nlidb}s}
\newcommand{\topl}{{\sc Top}}

\newcounter{phrasecounter}        

\newcommand{\phrase}[1]{{\small   
   \refstepcounter{phrasecounter}
   (\thephrasecounter) \ #1}}

\newcommand{\sphrase}[2]{{\small  
   \refstepcounter{phrasecounter} 
   (\thephrasecounter) #1#2}}

\newcommand{\samephrase}[1]{{\small 
   \hspace*{1.4 \parindent} #1}}

\newcommand{\lexpr}[1]{{\small    
   \refstepcounter{phrasecounter}
   (\thephrasecounter) \ $#1$}}

\newcommand{\samelexpr}[1]{{\small
   \hspace*{1.4 \parindent} $#1$}}

\newcommand{\phrases}[1]{         
   \smallskip
   \hspace*{-1.5\parindent}
   \begin{tabular}{l}
   #1
   \end{tabular}
   \smallskip}

\newcommand{\pref}[1]{(\ref{#1})} 

\documentstyle[acsc,twocolumn,epsf]{article}
\begin{document}
\bibliographystyle{acsc}

\raggedbottom
\def\BibTeX{{\rm B\kern-.05em{\sc i\kern-.025em b}\kern-.08em
    T\kern-.1667em\lower.7ex\hbox{E}\kern-.125emX}}

\title{A Framework for Natural Language Interfaces \\to Temporal Databases}
\author{{\em Ion Androutsopoulos}\\[1ex]
        Microsoft Research Institute\\ 
        Macquarie University \\ Sydney, NSW 2109, Australia \\[1ex]
        {\em ion@mri.mq.edu.au} \vspace*{4mm} \and
        {\em Graeme D.\ Ritchie}\\[1ex]
        Department of Artificial Intelligence \\ 
        University of Edinburgh\\ 
        80 South Bridge, Edinburgh EH1 1HN, Scotland, U.K. \\[1ex]
        {\em G.D.Ritchie@ed.ac.uk} \vspace*{4mm} \and
        {\em Peter Thanisch}\\[1ex]
        Department of Computer Science\\ 
        University of Edinburgh\\ 
        King's Buildings, Mayfield Road, Edinburgh EH9 3JZ, 
        Scotland, U.K.\\[1ex]
        {\em pt@dcs.ed.ac.uk}}
\date{}

\maketitle
\thispagestyle{empty}
        \begin{figure}[b]
        ~\\
        \noindent
        {\small\bf\raggedright
        \rule{3.5cm}{0.3mm} \hfill \\
        Proceedings of the 20th Australasian
        Computer Science Conference, Sydney, Australia,
        February 5--7 1997.
        }
        \end{figure}

\subsection*{\centering Abstract}
\noindent
{\it 
    Over the past thirty years, there has been considerable
    progress in the design of natural language interfaces to
    databases. Most of this work has concerned {\em snapshot}\/
    databases, in which there are only limited facilities for
    manipulating time-varying information. The database community is
    becoming increasingly interested in {\em temporal} databases,
    data-bases with special support for time-dependent entries. We 
    have developed a framework for constructing natural language
    interfaces to temporal databases, drawing on research on temporal
    phenomena within logic and linguistics. The central part of our
    framework is a logic-like formal language, called TOP, which
    can capture the semantics of a wide range of English sentences.
    We have implemented an HPSG-based sentence analyser that
    converts a large set of English queries involving time into 
    TOP formulae, and have formulated a provably correct procedure
    for translating TOP expressions into queries in the
    TSQL2 temporal data-base language. In this way we have established a
    sound route from English to a general-purpose temporal database
    language. 
}

\paragraph{Keywords} 
Natural language processing, natural language interfaces, temporal
databases. 

\section{Background} \label{introduction}

Time is an important research topic in both linguistics (tense and
aspect theories; see \cite{Comrie}, \cite{Comrie2} for an
introduction), and logic (temporal logics; see
\cite{VanBenthem}). Computer scientists are also becoming increasingly
interested in temporal databases, databases that are intended to store not only
present but also past and future facts, and that generally provide
special support for the notion of time \cite{Jensen} \cite{Tansel3}.
Although interesting ideas have emerged in all
three time-related disciplines, these ideas have remained largely
unexploited in the area of natural language interfaces to databases
(\nlidbs; see \cite{Perrault}, \cite{Copestake}, and
\cite{Androutsopoulos1995} for an introduction to
\nlidbs). Most \nlidbs\ cannot answer questions involving time,
because: (a) they cannot cope with the semantics of natural language
temporal expressions (e.g.\ verb tenses, temporal adverbials),
and (b) they were designed to interface to ``snapshot'' database
systems, that provide no special support for the notion of time. 

Previous research on \nlidbs\ for temporal data-bases has ignored
important temporal linguistic phenomena, used not fully defined
meaning representation languages, or assumed ad hoc temporal data-base
models and languages.  Clifford \cite{Clifford}, for example, has
defined formally a temporal version of the relational database model,
and a fragment of English that can be used to query databases
structured according to his database model. Clifford's approach is
interesting in that both the semantics of the English fragment and of
the temporal database model are defined within a Montague semantics
framework \cite{Dowty}. However, Clifford's coverage of English is extremely
narrow, and the semantics of the English mechanisms for expressing
time are oversimplified.  For example, perfect and continuous tenses
are not supported, and no distinction between states, events,
culminated activities, and points (section \ref{linguistic_data}
below) is made.  Furthermore, there is no indication that the overall
theory has ever been used to implement an actual \nlidb.

De et al.\ \cite{De} also support only an extremely limited subset of
English temporal mechanisms, and the underlying ``temporal database''
looks more like a collection of if-then-else rules than a principled
temporal database system. In the {\sc Cle} system \cite{Alshawi},
verb tenses introduce temporal operators (section \ref{logic}
below) and event/state variables into the generated logical
expressions. The semantics of these operators and the semantics of the
event/state variables, however, are left undefined. 

\smallskip

Past work on \nlidbs\ has shown the benefits of using a principled
intermediate representation language (typically, some form of logic)
to encode the meanings of natural language queries, with the resulting
intermediate language expressions being available for translation into
a suitable database language (e.g.\ {\sc Sql} \cite{Melton1993}).
Similar advantages (such as generality, modularity and portability;
see sections 5.4 and 6 of \cite{Androutsopoulos1995}) accrue from
developing temporal variants of this architecture. We have developed a
formal language, called {\sc Top}, to serve as the intermediate
representation language in place of conventional (non-temporal)
logics. A temporal extension of {\sc Sql}, called {\sc Tsql2}
\cite{TSQL2book}, was also proposed recently. Our architecture (in
direct reflection of existing \nlidbs) has an English query parsed
into a syntactic structure and converted into a {\sc Top} expression
encoding the relevant aspects of its meaning. This is then translated
into a {\sc Tsql2} query, and the evaluation of this query
against the temporal database supplies the answer to the original
English query. 

More specifically, we have addressed the following issues:
(a) design and implementation of a non-trivial English grammar
  handling temporal phenomena;
(b) design of the {\sc Top} language, including the definition of a
  precise model-theoretic semantics for it;
(c) devising a systematic conversion from English syntactic form to
    {\sc Top} formulae;
(d) defining translation rules from {\sc Top} formulae to {\sc
    Tsql2} queries, and proving the correctness of the translation rules;
(e) implementing all the above.
The full details of {\sc Top} and the translation to {\sc Tsql2} are
highly formal and rather voluminous, so such technical details are
beyond the scope of this paper. Here we concentrate on giving an
overview of the work and the motivation for some of the directions we
have followed.

Section \ref{linguistic_data} below surveys, from the perspective of
\nlidbs, some of the linguistic phenomena relating to
temporal information. This discussion demonstrates that there are real
linguistic issues involved in providing correct replies to English
queries directed to a temporal database.  Section \ref{logic} outlines
{\sc Top}, showing how it captures important semantic distinctions
that occur within English temporal queries.  Section
\ref{english_to_top} sketches how English sentences can be converted
systematically to {\sc Top} expressions, and section
\ref{top_to_tsql2} summarises the salient features of the translation
from {\sc Top} to {\sc Tsql2}.  We conclude with some remarks about
the direction which this kind of research could take in the future.

\section{The linguistic data} \label{linguistic_data}

There is a wealth of mechanisms for expressing time in English (and most
natural languages). Temporal information can be conveyed by verb
tenses, nouns (\qit{day}, \qit{beginning}), adjectives (\qit{earliest},
\qit{annual}), adverbs (\qit{yesterday}, \qit{twice}),
prepositional phrases (\qit{at 5:00pm}, \qit{for two hours}), and
subordinate clauses (\qit{while gate 2 was open}), to mention just
some of the temporal mechanisms. It is well-known that the semantics
of English temporal expressions cannot be modelled adequately in the
absence of some classification of verbs in terms of the situations
described by the verbs. (We use ``situation'' to refer collectively to
what other authors call ``event'', ``state'', ``action'', ``process'',
etc.) Most of the classifications that have been proposed originate
from Vendler's taxonomy \cite{Vendler}.  We use a version
of Vendler's taxonomy, whereby verbs are divided into: {\em state
verbs}, {\em activity verbs}, {\em culminated activity verbs}, and
{\em point verbs}.

Roughly speaking, state verbs describe a property
without referring to an action or a change in the world.  For example,
\qit{to contain} and \qit{to border}, as in \qit{Tank 2
  contains oil.} and \qit{Greece borders Bulgaria.}, are state verbs.
Activity verbs, in contrast, refer to actions or changes in the world.
\qit{To run} and \qit{to advertise}, as in \qit{John ran.} and
\qit{IBI advertised a new computer.}, are examples of activity verbs.
Culminated activity verbs are similar to activity verbs, in that they
describe world changes or actions. They differ, however, from activity
verbs in that the situations they describe have an inherent climax,
a point that has to be reached for the action/change to
be considered complete. \qit{To fix (an engine)} and \qit{to build (a
  bridge)}, as in \qit{Engineer 1 fixed engine 2.} and \qit{Housecorp
  built a bridge.}, are culminated activity verbs.  The climax of the
fixing is the point where the repair of the engine is finished,
and the climax of the building is the point where the 
construction of the bridge is completed.  In contrast, the
situations described by \qit{to run} and \qit{to advertise}
in \qit{John ran.} and \qit{IBI advertised a new computer.}
do not seem to have inherent climaxes. Finally, point verbs describe
situations that are perceived as instantaneous. \qit{To explode}, as in
\qit{A bomb exploded.}, is a point verb. 

The class of a verb may depend on the syntactic complements of the
verb (e.g.\ its object). For example, \qit{to run} with no object (as
in \qit{John ran.}) is an activity verb, but \qit{to run} with an
object denoting a specific distance (as in \qit{John ran a mile.}) is
a culminated activity verb (the climax is the point where John
completes the mile). Aspectual markers (e.g.\ the progressive
aspect) may cause a verb to be moved from its normal class to another
one (this will be discussed below).

The distinction between activity and culminated activity verbs can be
used to account for the so-called ``imperfective paradox''
\cite{Dowty1977} \cite{Lascarides}. 

\phrases{
\phrase{Was IBI ever advertising a new computer? \label{f1}}\\
\phrase{Did IBI ever advertise a new computer? \label{f2}}\\
\phrase{Was engineer 1 ever fixing engine 2? \label{f3}}\\
\phrase{Did engineer 1 ever fix engine 2? \label{f4}}
}

\noind If the \nlidb's answer to \pref{f1} is affirmative, then the
answer to \pref{f2} must also be affirmative. In contrast, if the
answer to \pref{f3} is affirmative, this does not necessarily imply
that the answer to \pref{f4} will also be affirmative (engineer 1 may
have abandoned the repair before completing it; we classify 
\qit{to advertise} as an activity verb, while \qit{to fix (an engine)}
as a culminated activity verb). In the case of culminated activity
verbs, the simple past (\qit{did fix}) requires the climax to have
been reached (i.e.\ the repair must have been completed).  In
contrast, the past continuous of culminated activity verbs (\qit{was
fixing}) makes no claim that the climax was reached.  Hence, an
affirmative answer to \pref{f3} does not imply an affirmative answer
to \pref{f4} (though an affirmative answer to \pref{f4} implies an
affirmative answer to \pref{f3}). In the case of activity verbs, there
is no climax, and neither the simple past nor the past continuous make
any claim that a climax was reached.  Hence, an affirmative answer to
\pref{f1} implies an affirmative answer to \pref{f2} (and vice versa).

The need for a classification of verbs is also apparent when verbs
combine with temporal adverbials (see also the linguistic data of
\cite{Kent}). When state verbs combine with adverbials understood as specifying
time points, the situation of the verb must usually simply hold at the
point of the adverbial. For example, in \pref{f5} any tank that
contained oil at 5:00pm must be reported. There is no requirement that
5:00pm must have been the point at which the tank started or stopped
containing oil. 

\phrases{
\phrase{Which tanks contained oil at 5:00pm? \label{f5}}\\
\phrase{Which athlete ran at 5:00pm? \label{f6}}\\
\phrase{Who fixed an engine at 5:00pm? \label{f7}}\\
\phrase{Which station broadcast the President's \label{f7a}}\\
\samephrase{message at 5:00pm?}
}

\noind In contrast, in the case of activity verbs, the \qit{at} point
is usually understood as the time at which the activity started. For
example, in \pref{f6}, the most natural reading is that the athlete
{\em started}\/ to run at 5:00pm. (In the progressive \qit{Which
  athlete was running at 5:00pm.}, however, the adverbial does not
have an inchoative meaning. This will be discussed below.) Finally,
with culminated activity verbs (in non-progressive forms), the
\qit{at} point is usually the time at which the climax was reached, or
in some cases the point where the change/action described by the verb
started. In \pref{f7}, for example, 5:00pm is probably the time at
which the repair was completed. The inchoative meaning with
culminated activity verbs is easier to accept in \pref{f7a}, where
5:00pm is probably the point where the broadcasting started. (We
classify \qit{to broadcast (a message)} as a culminated activity verb,
with the climax being the point where the broadcasting of the message
is completed.)

Verb aspects also play an important role. In \pref{f11b}, the most
natural reading is that 5:00pm must simply have been a point where the
running was ongoing.  There is no implication that the running must
have started at 5:00pm. (The futurate meanings of progressive tenses
-- e.g.\ the athlete in \pref{f11b} was {\em going to} run at 5:00pm,
but perhaps never ran -- are ignored in this project.)  Compare
\pref{f11b} to \pref{f6}, where 5:00pm is probably the time where the
running started.

\phrases{
\phrase{\ Which athlete was running at 5:00pm? \label{f11b}}\\
\phrase{Who was fixing an engine at 5:00pm? \label{f12}}\\
\phrase{Which station was broadcasting the \label{f13}} \\
\samephrase{\hspace*{0.4mm} President's message at 5:00pm?}
}

\noind In other words, although \qit{to run} is an activity verb, in
\pref{f11b} it behaves as if it were a state verb. (With 
state verbs, the adverbial's point is simply a point where the
situation was true.) Similar observations can be made for
\pref{f12} and \pref{f13} (cf.\ \pref{f7} and \pref{f7a}).
We account for \pref{f11b}--\pref{f13} by assuming that the
progressive verb aspect transforms activity and culminated activity verbs
into state verbs. (This is similar to Moens' view 
\cite{Moens} that the progressive coerces
``processes'' into states.) 

A cancelling transformation (see section \ref{english_to_top} below)
takes place when culminated activity verbs combine with \qit{for}
adverbials. This transformation cancels the normal implication 
that the climax has been reached. For
example, \pref{f18} implies that the climax has been reached. In
contrast, \pref{f19} carries no such implication.

\phrases{
\phrase{Housecorp built bridge 2. \label{f18}}\\
\sphrase{?}{Housecorp built bridge 2 for two years. \label{f19}}\\
\phrase{Housecorp was building bridge 2 for two years.\label{f20}}\\
\sphrase{*}{John fixed fault 2 for two hours. \label{f21}}\\
\phrase{John was fixing fault 2 for two hours. \label{f22}}
}

\noind Some native speakers find \pref{f19} unacceptable, and
\pref{f21} is unacceptable to most native speakers. It seems, however,
a reasonable simplification to assume that a \nlidb\ could treat
\pref{f19} and \pref{f21} as grammatical, and equivalent to \pref{f20}
and \pref{f22} respectively. (In \pref{f20} and \pref{f22} there is no
implication that a climax was reached.) 

\smallskip

We note at this point that we have focused our work on stand-alone
questions. We have not examined discourse-related phenomena
\cite{Kamp1993}. We have also restricted our work to
questions about the past and the present. We have not examined
questions referring to the future.

\section{Modelling time in TOP} \label{logic}

This section provides an overview of \topl, the formal language we use
to represent the meanings of the English questions. \topl\ assumes
that time is linear, discrete, and bounded \cite{VanBenthem}, and expresses
temporal information using operators (\topl\ stands for
``language with Temporal OPerators''). For example, \pref{f39} would
be expressed in \topl\ as \pref{f40}:

\phrases{
\phrase{Did tank 2 (ever) contain water? \label{f39}}\\
\lexpr{\opr{Past}[contain(tank2, water)] \label{f40}}
}

\noind where $\opr{Past}$\/ is a temporal operator, which roughly
speaking requires $contain(tank2, water)$\/ to be true at some past
time. The answer to \pref{f39} is affirmative if and only if
\pref{f40} evaluates to true.

\topl's temporal operators have been influenced by those 
of \cite{Crouch2}. An alternative operator-less approach would be
to introduce time as an extra argument of each predicate. In this
case, \pref{f39} would be expressed as:

\phrases{
\lexpr{\exists t' \; contain(tank2, water, t') \land t'<now \label{f41}}
}

\noind where $<$\/ denotes temporal precedence. (In this and following
sections primed strings are used as variables.)
We use temporal operators mainly because they lead to more compact
formulae. We make no claim regarding the expressivity of \topl\ and
other operator-based languages vs.\ operator-less languages.  

\subsubsection*{Speech, event, and localisation time}

{\sc Top} formulae are evaluated with respect to three parameters:
{\em speech time} ($st$\/), {\em event time} ($et$\/), and {\em
localisation time} ($lt$\/). The first two are as in Reichenbach's
work \cite{Reichenbach}.  $st$\/ is the time point where the question is
submitted to the \nlidb. $et$\/ is, roughly speaking, an interval
corresponding to the time where the situation represented by the
formula takes place. The third parameter, $lt$, derives from the logic
of \cite{Crouch2}. (It has nothing to do with Reichenbach's ``reference
time''.) $lt$\/ is an interval acting as a
temporal window within which $et$ must be located. 

To understand how the three parameters work, let us consider the reading of
\pref{f41.1} that asks if John was running some time on
1/6/94. The corresponding \topl\ formula is \pref{f41.2}.

\phrases{
\phrase{Did John run on 1/6/94? \label{f41.1}}\\
\lexpr{\opr{At}[1/6/94, \opr{Past}[run(john)]] \label{f41.2} }
}

\noind \pref{f41.2} is evaluated as follows. First, $st$\/ is fixed to
the point where \pref{f41.1} was submitted to the \nlidb.  Initially,
$lt$\/ covers the whole time-axis, and $et$\/ can be any
interval. Next, the $\opr{At}$\/ operator narrows the localisation
time window, so that it only covers the day 1/6/94. Thus, $et$\/ now
has to be a subinterval of 1/6/94. The $\opr{Past}$\/ operator
(introduced by the verb tense) requires $lt$\/ to be narrowed, so that
it only contains time points that precede $st$. (If the question is
submitted after 1/6/94, the $\opr{Past}$ does not narrow $lt$\/ any
further.) \pref{f41.2} evaluates to true, if and only if it is
possible to find an $et$\/ where $run(john)$\/ is true (i.e.\ John was
running throughout that interval), such that $et$\/ is a subinterval
of $lt$\/ (i.e.\ a subinterval of 1/6/94, if the question is submitted
after 1/6/94).

\subsubsection*{Homogeneity}

{\sc Top} atomic formulae (predicates) always satisfy the following
{\em homogeneity restriction}\/: if an atomic formula (e.g.\
$contains(tank2, water)$\/) is true at an event time $et_1$, then it
is also true at any event time $et_2$ that is a subinterval of
$et_1$. Non-atomic {\sc Top} formulae do not have to satisfy this
restriction. (Various versions of homogeneity have been used in
\cite{Allen1984}, \cite{Richards},
\cite{Kent}, and elsewhere.)

\subsubsection*{Progressives}

The progressives of activity and point verbs are expressed using the same
predicates that express the corresponding non-progressive forms. For
example, the reading of \pref{f46} that asks if John was
running at some time on 1/6/94 is expressed using \pref{f41.2}, the
same \topl\ formula that expresses the non-progressive \pref{f41.1}.

\phrases{
\phrase{Was John running on 1/6/94? \label{f46}}\\
}

\noind Progressives of culminated activity verbs are expressed in a similar
manner. For example, the reading of \pref{f48} that asks
if John was fixing engine 2 some time on 1/6/94 is
expressed as \pref{f49}. 

\phrases{
\phrase{Was John fixing engine 2 on 1/6/94? \label{f48}}\\
\lexpr{\opr{At}[\mtxt{1/6/94}, \opr{Past}[fixing(john, eng2)]] \label{f49}} 
}

\noind State verbs typically do not appear in
progressive forms (e.g.\ \qit{Tank 2 was containing water.} sounds odd). 

\subsubsection*{Non-progressives of culminated activity verbs}

Non-progressive forms of culminated activity verbs
are expressed using the $\opr{Culm}$\/ operator and the predicates that
correspond to the progressive forms. For example, \pref{f50} is expressed
as \pref{f51}.

\phrases{
\phrase{Did John fix engine 2 on 1/6/94? \label{f50}}\\
\lexpr{\opr{At}[\mtxt{1/6/94}, \opr{Past}[\opr{Culm}[fixing(john, eng2)]]]
 \label{f51}}
}

\noindent In \pref{f51}, the semantics of the $\opr{Culm}$ operator
requires $et$ to cover a maximal interval where the predicate
$fixing(john, eng2)$ is true, the end-point of $et$\/ to be a point
where the repair reaches its climax, and $et$\/ to be a subinterval of
$lt$.  Assuming that \pref{f50} is submitted after 1/6/94, when the
expression $\opr{Culm}[fixing(john, eng2)]$\/ is evaluated, $lt$\/ is the
interval that covers exactly the day 1/6/94. The answer to \pref{f50}
will be affirmative if and only if for some event time interval $et$,
$et$ covers exactly a repair (from start to completion) of engine 2 by
John, and $et$ is a subinterval of 1/6/94.

\pref{f51} captures the reading of \pref{f50} whereby a
 repair of engine 2 by John must have both started and been completed
within 1/6/94. Under an alternative reading, it is enough if the
repair simply reached its climax on 1/6/94. In this case, the repair
may have started, for example, the day before. This reading is
captured by \pref{f52}.

\phrases{
\lexpr{\opr{At}[\mtxt{1/6/94},\label{f52}}\\
\samelexpr{\;\opr{Past}[\opr{End}[\opr{Culm}[fixing(john, eng2)]]]]}
}

\noind According to \pref{f52}, it is enough if the {\em end-point}\/ of
an interval that covers exactly a repair from start to completion
falls within 1/6/94.

An affirmative answer to \pref{f48} (expressed as \pref{f49}) does not
necessarily imply an affirmative answer to \pref{f50} (expressed as
\pref{f51}). If, for example, John was fixing engine 2 some time on
1/6/94, but never completed the repair, then there will be an interval
within 1/6/94 at which $fixing(john, eng2)$\/ is true, but there
will be no interval at which the expression $\opr{Culm}[fixing(john, eng2)]$\/
is true, because at no point did the repair reach its climax.  Hence, the
answer to \pref{f49} will be affirmative, but the answer to \pref{f51}
will be negative. This accords with the imperfective paradox of
section \ref{linguistic_data}. (It should also be easy to see that an
affirmative answer to \pref{f51} implies an affirmative answer to
\pref{f49}.)

\subsubsection*{Wh-questions}

So far, we have considered only yes/no questions. Questions like
\pref{f53} are expressed using the interrogative quantifier $?$, as
shown in \pref{f54}.

\phrases{
\phrase{What did John fix? \label{f53}}\\
\lexpr{?x' \; \opr{Past}[\opr{Culm}[fixing(john,x')]] \label{f54}}
}

\noind \pref{f54} says that the answer should contain any $x'$,
such that John completed the fixing of $x'$\/ in the past. 

The $\opr{Past}$ operator actually has a slightly more complex form
than the one we have been using up to this point: it is indexed by a
variable ($e'$\/ in the following example). The \topl\ formula for
\pref{f53} would actually be \pref{f55}.

\phrases{
\lexpr{?x' \; \opr{Past}[e', \opr{Culm}[fixing(john,x')]] \label{f55}}
}

\noind The semantics of \topl\ binds $e'$\/ to $et$. The
$e'$\/ variable of $\opr{Past}$ is useful in time-asking questions like
\pref{f56}, expressed as \pref{f57}.

\phrases{
\phrase{When did tank 2 contain water? \label{f56}}\\
\lexpr{?_{mxl} e' \; \opr{Past}[e', contain(tank2, water)] \label{f57}}
}

\noindent \pref{f57} reports the maximal intervals among the past
intervals at which tank 2 contained water.

\subsubsection*{Perfective aspect}

\begin{figure}
\hrule
\begin{center}
\mbox{\epsfysize=1.6cm \epsffile{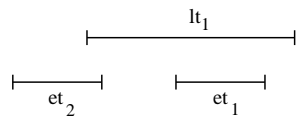}}
\caption{The $\opr{Perf}$ operator}
\label{perf_fig}
\end{center}
\hrule
\end{figure}

The perfective aspect is expressed using a special $\opr{Perf}$
operator. Ignoring some details, $\opr{Perf}[e_2',
\phi]$ is true with respect to a speech time $st$, an event time
$et_1$, and a localisation time $lt_1$, if and only if (see figure
\ref{perf_fig}): (a) $et_1$\/ is a subinterval of $lt_1$, (b) there is an
$et_2$\/ that ends before $et_1$, and (c) $\phi$\/ is true with
respect to $st$, $et_2$, and $lt_2$, where $lt_2$\/ covers the entire
time axis (i.e.\/ $lt$\/ is reset to the whole time axis when
evaluating $\phi$\/). $e2'$\/ is similar to the indexing variable of
$\opr{Past}$: $e2'$\/ is always bound to $et_2$. Intuitively,
$\opr{Perf}[e_2', \phi]$ is true at event time intervals that are
preceded by other event time intervals where $\phi$\/ is true.  To
illustrate the use of $\opr{Perf}$, let us consider \pref{f58}.

\phrases{
\phrase{Had IBI advertised PPC on 1/1/85? \label{f58}}\\
\phrase{Did IBI advertise PPC on 1/1/85? \label{f59}}
}

\noind \pref{f58} has two readings. Under the first reading, it 
asks if IBI advertised PPC on 1/1/85 (remote past meaning). In this
case, \pref{f58} is similar to \pref{f59}. Under a second reading,
\pref{f58} asks if IBI had ever advertised PPC at any time up to (and
possibly including) 1/1/85.  Under the latter reading, if IBI
advertised PPC only on 6/6/84, the answer to \pref{f58} would still be
affirmative. The two readings are captured by \pref{f60} and
\pref{f61} respectively (our system generates both).

\phrases{
\lexpr{\opr{Past}[e_1', \opr{Perf}[e_2', \opr{At}[1/1/85,
  advertise(ibi,ppc)]]] \label{f60}}\\
\lexpr{\opr{At}[1/1/85, \opr{Past}[e_1', \opr{Perf}[e_2',
  advertise(ibi,ppc)]]] \label{f61}}
}

\noind Intuitively, \pref{f60} says that there must be a past event
time interval $e_1' = et_1$, that is preceded by another event time
interval $e_2' = et_2$, such that $e_2'$\/ falls within 1/1/85, and
$advertise(ibi,ppc)$\/ is true at $e_2'$. In other words, the
advertising takes place on 1/1/85.  In contrast, \pref{f61} says that
there must be a past event time interval $e_1' = et_1$, that falls
within 1/1/85, and that is preceded by another event time interval
$e_2' = et_2$\/ where $advertise(ibi,ppc)$\/ is true. In this case,
the advertising does not necessarily take place on 1/1/85.

\smallskip

We should point out that we have examined only the following tenses:
simple present, simple past, present continuous, past continuous,
present perfect, and past perfect. We have not examined how other
tenses could be expressed in {\sc Top}. Also, we have specified how to
express in {\sc Top} temporal subordinate clauses introduced by only
\qit{while}, \qit{before}, and \qit{after} (e.g.\ we have not
considered clauses introduced by \qit{when} or \qit{since}). Finally,
we have not examined how to express in {\sc Top} temporal adjectives
(e.g.\ \qit{first}, \qit{annual}), nouns introducing events (e.g.\
\qit{the construction of bridge 2}), order nouns (e.g.
\qit{predecessor}), or frequency adverbials (e.g.\ \qit{twice}).

\section{From English to TOP} \label{english_to_top}

The English questions are parsed and mapped to \topl\ expressions
using an {\sc Hpsg}-based grammar \cite{Pollard2}. The grammar was
developed using {\sc Ale} \cite{Carpenter1994}, and it is
based on previous {\sc Ale} encodings of {\sc Hpsg} fragments by
Penn, Carpenter, Manandhar, and Grover. Our grammar is very close to
the {\sc Hpsg} version of chapter 9 of \cite{Pollard2}, with the main
exception being that the situation theoretic semantic constructs of
\cite{Pollard2} have been replaced by feature structures that
represent {\sc Top} expressions. A detailed description of our grammar
is outside the scope of this paper (see
\cite{Androutsopoulos1996}). Here we will only attempt to offer a
flavour of how the grammar works.

Let us consider \pref{f80}, which our experimental system treats as a
yes/no question.

\phrases{
\phrase{John fixed an engine on 1/6/94. \label{f80}}
}

\begin{figure}
\hrule
\begin{center}
\mbox{\epsfysize=3.3cm \epsffile{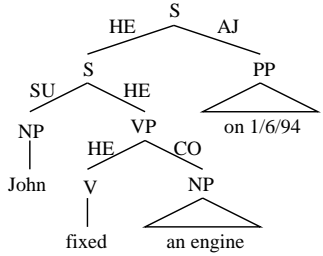}}
\caption{Parse tree for \qit{John fixed an engine on 1/6/94.}}
\label{parse_tree_fig}
\end{center}
\hrule
\end{figure}

\noind Figure \ref{parse_tree_fig} shows the parse tree for
\pref{f80}. Arcs marked with {\sc he}, {\sc su}, {\sc co}, and {\sc
aj} correspond to head, subject, complement, and adjunct daughters
respectively.  The lexical head (the verb \qit{fixed}) first combines
with its complement (the noun phrase \qit{an engine}). The
resulting verb phrase combines with its subject (\qit{John}),
producing a sentence. The prepositional phrase \qit{on 1/6/94}
attaches to this sentence as an adjunct. In our grammar, temporal adjuncts like
\qit{on 1/6/94} or \qit{yesterday} are taken to modify full sentences
(verbs that have combined with their subjects and complements). There
is only one case where our grammar allows temporal adjuncts to modify
verb phrases (verbs that have combined with their complements but not
their subjects), and this is in the case of past participles (e.g.\
\qit{given}). Unlike all other verb forms, we allow past participles
to be modified by temporal adverbials either before or after combining
with their subjects. This is needed to be able to generate both
readings of \pref{f58}. 

In {\sc Hpsg}, the order in which the daughters of a node appear in the
surface sentence is determined by the Constituent Ordering Principle
({\sc Cop}). Our version of {\sc Cop} places no restriction on
the order between temporal adjuncts like \qit{on 1/6/94} and the head
daughters that the adjuncts modify. Hence, \qit{on 1/6/94} can either follow
the \qit{John assembled an engine} as in \pref{f80}, or it can precede
it as in \pref{f82}. In either case, the \topl\ formula would be the
same, i.e.\ \pref{f82a}.

\phrases{
\phrase{On 1/6/94 John fixed an engine. \label{f82}} \\
\lexpr{\exists x' \; engine(x') \; \land  \label{f82a}} \\
\samelexpr{\opr{At}[1/6/94, \opr{Past}[e',\opr{Culm}[fixing(john, x')]]]}
}

Let us now examine how \pref{f80} (or \pref{f82}) is mapped to
\pref{f82a}. A lexicon entry associates the past tense form \qit{fixed}
with the expression in \pref{f84}.\footnote{In our system, the person
that configures the lexicon needs to provide lexicon entries for only
the base forms of verbs. Lexicon entries for non-base verb forms are
generated automatically by lexical rules.} (``$\_\_$'' denotes an
empty slot.)

\phrases{
\lexpr{\opr{Past}[e',\opr{Culm}[fixing(\_\_,\_\_)]] \label{f84}}
}

The noun phrase \qit{an engine} receives the
expression shown in \pref{f85}. (The existential quantifier derives
from the determiner \qit{a}, and the $engine(x')$\/ derives from the
lexical entry for \qit{engine}.)

\phrases{
\lexpr{\exists x' \; engine(x') \label{f85}}
}

When \qit{fixed} combines with \qit{an engine}, \pref{f85} enters a
quantifier store \cite{Cooper1983}, and $x'$, the variable used in
\pref{f85}, fills the second argument-slot of the $fixing(\_\_,
\_\_)$\/ in \pref{f84}. The verb phrase\qit{fixed an engine} inherits
the semantics of its head daughter (\qit{fixed}), but now the second
argument of the predicate $fixing(\_\_, \_\_)$\/ is $x'$:

\phrases{
\lexpr{\opr{Past}[e',\opr{Culm}[fixing(\_\_, x')]] \label{f86}}
}

Ignoring some details, when the verb phrase combines with its subject,
the constant corresponding to \qit{John} fills the remaining empty
slot of the predicate $fixing(\_\_, x')$, and the mother of the verb phrase
inherits the semantics of its head daughter, which is now:

\phrases{
\lexpr{\opr{Past}[e',\opr{Culm}[fixing(john, x')]] \label{f87}}
}

Finally, \qit{on 1/6/94} is mapped to \pref{f88}. When \qit{on 1/6/94}
combines with \qit{John assembled an engine}, the empty slot of
\pref{f88} is filled by the expression of the head daughter, i.e.\
\pref{f87}. Thus, \pref{f88} becomes \pref{f89}.

\phrases{
\lexpr{\opr{At}[1/6/94, \_\_] \label{f88}} \\
\lexpr{\opr{At}[1/6/94, \opr{Past}[e',\opr{Culm}[fixing(john, x')]]] 
   \label{f89}}  
}

\pref{f82a} is then generated by ``unstoring'' the contents of the
quantifier store in front of \pref{f89}, i.e.\ by adding \pref{f85} in
front of \pref{f89}. In our experimental system the unstoring
operation is quite primitive. The contents of the quantifier store are
simply added in front of the matrix expression, preserving the order
in which the quantifiers appear in the sentence. More elaborate
unstoring techniques are possible (see chapter 8 of \cite{Alshawi}). 

In a similar way, \pref{f93} is mapped to \pref{f94}. In this case,
a lexicon entry associates the present participle \qit{fixing} with
$fixing(\_\_,\_\_)$. The $\opr{Past}$ operator in \pref{f94} is
added by the auxiliary \qit{was}.

\phrases{
\phrase{John was fixing an engine on 1/6/94. \label{f93}}\\
\lexpr{\exists x' \; engine(x') \; \land \label{f94}} \\
\samelexpr{\opr{At}[1/6/94, \opr{Past}[e',fixing(john, x')]]}
}

The transformation that cancels the implication of culminated activity
verbs that the climax has been reached when a \qit{for} adverbial is
present (section \ref{linguistic_data}) has been implemented as a
post-processing rule. \pref{f104} is initially mapped to \pref{f105}.
(The $\opr{For}$ operator specifies the duration of the event time.)
The post-processing rule then removes any $\opr{Culm}$ operator from
the interior of a $\opr{For}$ operator that has been introduced by a
\qit{for~\dots} adverbial, resulting in \pref{f106}, the
same formula that expresses \pref{f104p}.

\phrases{
\phrase{Housecorp built bridge 2 for two years. \label{f104}}\\
\lexpr{\opr{For}[\mtxt{year}, 2, \label{f105}} \\
\samelexpr{\;\opr{Past}[e',\opr{Culm}[building(housecorp,bridge2)]]]}\\
\phrase{Housecorp was building bridge 2 for two years. \label{f104p}}\\
\lexpr{\opr{For}[\mtxt{year}, 2,\label{f106}} \\ 
\samelexpr{\;\opr{Past}[e',building(housecorp,bridge2)]]}
}

Interrogatives like \qit{who}, \qit{what}, or \qit{which engine} are
treated like normal noun phrases (e.g.\ \qit{an engineer}),
except that they insert interrogative quantifiers into the
quantifier store.  For example, \qit{which engineer} would cause $?x'
\; engineer(x')$\/ to be inserted into the quantifier
store. \pref{f109} is parsed in the same way as \pref{f108}, except
that the resulting formula contains an interrogative quantifier rather
than an existential one.  (Our system ignores punctuation.)

\phrases{
\phrase{Which engineer fixed an engine? \label{f109}}\\
\phrase{An engineer fixed an engine. \label{f108}}\\
}

\qit{When} is treated syntactically as a temporal adjunct, like
\qit{on 1/6/94} and \qit{yesterday}. \pref{f113} is analysed
syntactically in the same way as \pref{f114}. Unlike adjuncts like
\qit{on 1/6/94}, however, that introduce $\opr{At}$ operators,
\qit{when} introduces a $?_{mxl} e'$ (section \ref{logic}), where
$e'$\/ represents the event time.  The \topl\ formulae for \pref{f113}
and \pref{f114} are given in \pref{f113f} and \pref{f114f}
respectively.

\phrases{
\phrase{When did IBI advertise PPC? \label{f113}}\\
\lexpr{?_{mxl} e' \; \opr{Past}[e', advertise(ibi, ppc)] \label{f113f}}\\
\phrase{On 1/6/94, did IBI advertise PPC? \label{f114}}\\
\lexpr{\opr{At}[1/6/94, \opr{Past}[e', advertise(ibi, ppc)]] \label{f114f}}
}

\vspace*{-2mm}

\section{From TOP to TSQL2} \label{top_to_tsql2}

As remarked in section \ref{introduction}, there are various
advantages to the traditional \nlidb\ architecture in which 
natural language queries are systematically translated into an
intermediate logical language, then transformed into expressions
in an established database language. 

For conventional (``snapshot'') relational data-bases, the de facto
standard query language is {\sc Sql} \cite{Melton1993}.  In the newer
field of temporal databases, the position is less clear.  More than a
dozen temporal extensions of the relational database model have been
proposed, and there are also several proposals on how to modify {\sc
Sql} to support the notion of time (see, for example,
\cite{McKenzie}). We have chosen to adopt {\sc Tsql2}
\cite{TSQL2book}, a temporal extension of {\sc Sql} that was recently
proposed by a group comprising most leading temporal database
researchers. We have also adopted the proposed conceptual database model
for {\sc Tsql2}, called {\sc Bcdm}, as a formal basis for reasoning
about the meaning of {\sc Tsql2} expressions, and we have assumed that
there are abstract functions which will evaluate a {\sc Tsql2}
expression with respect to an arbitrary {\sc Bcdm} database.

A number of modifications to the {\sc Tsql2} specification have been
made during this project.  These can be classed as follows: (i) There
are minor alterations to achieve uniformity or consistency in places
where the {\sc Tsql2} definition contains some slight
discrepancies. (ii) It could be argued that some extensions are
desirable to improve the expressive power of the language, regardless
of natural language issues. (iii) Some extra facilities have been
incorporated to reflect subtleties of meaning which are largely
motivated by the richness of the natural language input, and which
might not be felt necessary in a purely database context.
Nevertheless, all such alterations have been kept to the minimum, and
the resulting version of {\sc Tsql2} is very close to the original
language. As our system is a research prototype, we are confident that
these extensions do not undermine the usefulness of the experiment.

\begin{figure}
\hrule
\begin{center}
\mbox{\epsfysize=2.7cm \epsffile{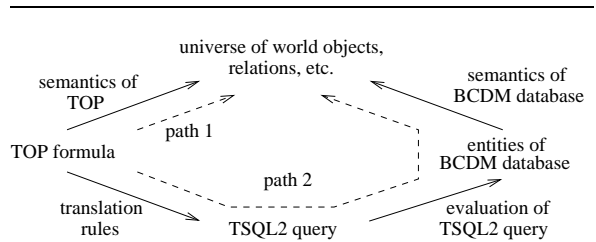}}
\caption{Proving the correctness of the translation from {\sc Top} to
  {\sc Tsql2}}
\label{translation_fig}
\end{center}
\hrule
\end{figure}

It is not feasible to provide here a complete account of our formal
method of translation and its correctness, but an outline of the
approach is possible.  We assume that a {\sc Top} expression and a
{\sc Tsql2} query both refer to some universe of world objects,
relations, etc.\ (much as in first-order logic), including temporal
entities (such as intervals). The denotation, in terms of this
universe, of a {\sc Top} formula is provided by the semantic
definitions which we have provided for the language (see
\cite{Androutsopoulos1996}). The denotation of a {\sc Tsql2}
expression is given indirectly, in that we assume that there is an
``evaluation'' function which will map a {\sc Tsql2} query to some
entities within a {\sc Bcdm} database, and that the semantics of the
{\sc Bcdm} database indicates how such database entities are related
to the universe of world entities, relations, time-intervals, etc. The
situation is roughly as in figure
\ref{translation_fig}.

There are a number of translation rules (implemented in Prolog) for
converting {\sc Top} expressions into {\sc Tsql2}. These are defined
recursively, in terms of a few basic types of {\sc Top} expressions
and of combinations of expressions. We have proven that the rules are
correct, in the sense that the denotation of a {\sc Top} query
(roughly speaking, its answer) as defined by the semantics of {\sc
Top} is the same as the denotation (answer) of the corresponding {\sc
Tsql2} query when determined by way of the ``evaluation'' function and
the {\sc Bcdm} database semantics (see \cite{Androutsopoulos1996} for
the complete proof). In terms of figure
\ref{translation_fig} above, the same semantic content is reached
whether path 1 or path 2 is chosen.

\section{Future directions}

As mentioned in section \ref{logic}, there are several kinds of
English temporal expressions for which we have not examined possible
representations in {\sc Top} (e.g.\ expressions referring to the
future, temporal adjectives, etc.). It would be interesting to explore
if our framework can be extended, so that questions containing these
expressions can also be mapped systematically to an (extended version
of) {\sc Top} and then to {\sc Tsql2}.

A major practical limitation of our prototype \nlidb\ is that it has
never been linked to an actual database management system ({\sc
Dbms}), mainly because until recently no {\sc Dbms} supported {\sc
Tsql2}. This means that the generated {\sc Tsql2} queries are not
executed, and no answers are produced. An experimental {\sc Dbms},
called {\sc TimeDb}, that supports {\sc Tsql2} is now available (see
\cite{Boehlen1995c}), and it would be interesting to attempt to link
our \nlidb\ to that {\sc Dbms}. This task is complicated by the fact
that both our framework and {\sc TimeDb} use different versions of
{\sc Tsql2}.

\section*{Acknowledgements}

This paper reports on work that was carried out while the first author
was in the Department of Artificial Intelligence, University of
Edinburgh, supported by the Greek State Scholarships Foundation.

\bibliography{biblio}

\begin{thebibliography}{10}

\bibitem{Allen1984}
J.F. Allen.
\newblock {Towards a General Theory of Action and Time}.
\newblock {\em Artificial Intelligence}, Volume~23, pages 123--154, 1984.

\bibitem{Alshawi}
H.~Alshawi (editor).
\newblock {\em {The Core Language Engine}}. MIT Press, Cambridge,
  Massachusetts, 1992.

\bibitem{Androutsopoulos1996}
I.~Androutsopoulos.
\newblock {\em {A Principled Framework for Constructing Natural Language
  Interfaces to Temporal Databases}}.
\newblock Ph.D.\ thesis, Department of Artificial Intelligence, University of
  Edinburgh, 1996.

\bibitem{Androutsopoulos1995}
I.~Androutsopoulos, G.D. Ritchie and P.~Thanisch.
\newblock {Natural Language Interfaces to Databases -- An Introduction}.
\newblock {\em Natural Language Engineering}, Volume~1, Number~1, pages 29--81,
  1995.

\bibitem{Boehlen1995c}
M.H. Boehlen.
\newblock {Temporal Database System Implementations}.
\newblock Unpublished document, Department of Mathematics and Computer Science,
  Aalborg University, 1995.

\bibitem{Carpenter1994}
B.~Carpenter and G.~Penn.
\newblock {The Attribute Logic Engine -- User's Guide}.
\newblock Unpublished document, Philosophy Department, Carnegie Mellon
  University, December 1994.

\bibitem{Clifford}
J.~Clifford.
\newblock {\em {Formal Semantics and Pragmatics for Natural Language
  Querying}}.
\newblock Cambridge Tracts in Theoretical Computer Science, Cambridge
  University Press, 1990.

\bibitem{Comrie}
B.~Comrie.
\newblock {\em {Aspect}}.
\newblock Cambridge University Press, 1976.

\bibitem{Comrie2}
B.~Comrie.
\newblock {\em {Tense}}.
\newblock Cambridge University Press, 1985.

\bibitem{Cooper1983}
R.~Cooper.
\newblock {\em {Quantification and Syntactic Theory}}.
\newblock D. Reidel, Dordrecht, Holland, 1983.

\bibitem{Copestake}
A.~Copestake and K.~Sparck~Jones.
\newblock {Natural Language Interfaces to Databases}.
\newblock {\em The Knowledge Engineering Review}, Volume~5, Number~4, pages
  225--249, 1990.

\bibitem{Crouch2}
R.S. Crouch and S.G. Pulman.
\newblock {Time and Modality in a Natural Language Interface to a Planning
  System}.
\newblock {\em Artificial Intelligence}, Volume~63, pages 265--304, 1993.

\bibitem{De}
S.~De, S.~Pan and A.B. Whinston.
\newblock {Natural Language Query Processing in a Temporal Database}.
\newblock {\em Data \& Knowledge Engineering}, Volume~1, pages 3--15, 1985.

\bibitem{Dowty1977}
D.R. Dowty.
\newblock {Toward a Semantic Analysis of Verb Aspect and the English
  `Imperfective' Progressive}.
\newblock {\em Linguistics and Philosophy}, Volume~1, pages 45--77, 1977.

\bibitem{Dowty}
D.R. Dowty, R.E. Wall and S.~Peters.
\newblock {\em {Introduction to Montague Semantics}}.
\newblock D.Reidel Publishing Company, Dordrecht, Holland, 1981.

\bibitem{Jensen}
C.S. Jensen, J.~Clifford, S.K. Gadia, A.~Segev and R.T. Snodgrass.
\newblock {A Glossary of Temporal Database Concepts}.
\newblock {\em SIGMOD Record}, Volume~21, Number~3, pages 35--43, 1992.

\bibitem{Kamp1993}
H.~Kamp and U.~Reyle.
\newblock {\em {From Discourse to Logic}}.
\newblock Kluer Academic Publishers, 1993.

\bibitem{Kent}
S.~Kent.
\newblock {\em {Modelling Events from Natural Language}}.
\newblock Ph.D.\ thesis, Imperial College of Science Technology and Medicine,
  1993.

\bibitem{Lascarides}
A.~Lascarides.
\newblock {\em {A Formal Semantic Analysis of the Progressive}}.
\newblock Ph.D.\ thesis, University of Edinburgh, 1988.

\bibitem{McKenzie}
E.~McKenzie and R.~Snodgrass.
\newblock {Evaluation of Relational Algebras Incorporating the Time Dimension
  in Databases}.
\newblock {\em ACM Computing Surveys}, Volume~23, Number~4, pages 501--543,
  1991.

\bibitem{Melton1993}
J.~Melton and A.R. Simon.
\newblock {\em {Understanding the New SQL: A Complete Guide}}.
\newblock Morgan Kaufmann Publishers, San Mateo, California, 1993.

\bibitem{Moens}
M.~Moens.
\newblock {\em {Tense, Aspect and Temporal Reference}}.
\newblock Ph.D.\ thesis, University of Edinburgh, 1987.

\bibitem{Perrault}
C.R. Perrault and B.J. Grosz.
\newblock {Natural Language Interfaces}.
\newblock In H.E. Shrobe (editor), {\em {Exploring Artificial Intelligence}},
  pages 133--172. Morgan Kaufmann Publishers Inc., San Mateo, California, 1988.

\bibitem{Pollard2}
C.~Pollard and I.A. Sag.
\newblock {\em {Head-Driven Phrase Structure Grammar}}.
\newblock University of Chicago Press and Center for the Study of Language and
  Information, Stanford., 1994.

\bibitem{Reichenbach}
H.~Reichenbach.
\newblock {\em {Elements of Symbolic Logic}}.
\newblock Collier-Macmillan, London, 1947.

\bibitem{Richards}
B.~Richards, I.~Bethke, J.~van~der Does and J.~Oberlander.
\newblock {\em {Temporal Representation and Inference}}.
\newblock Cognitive Science Series, Academic Press, Harcourt Brace Jovanovich
  Publishers, 1989.

\bibitem{TSQL2book}
R.T. Snodgrass (editor).
\newblock {\em {The TSQL2 Temporal Query Language}}. Kluwer Academic
  Publishers, 1995.

\bibitem{Tansel3}
A.~Tansel, J.~Clifford, S.K. Gadia, S.~Jajodia, A.~Segev and R.T. Snodgrass.
\newblock {\em {Temporal Databases -- Theory, Design, and Implementation}}.
\newblock Benjamin/Cummings, California, 1993.

\bibitem{VanBenthem}
J.F.A.K. van Benthem.
\newblock {\em {The Logic of Time}}.
\newblock D.Reidel Publishing Company, Dordrecht, Holland, 1983.

\bibitem{Vendler}
Z.~Vendler.
\newblock {Verbs and Times}.
\newblock In {\em Linguistics in Philosophy}, Chapter~4, pages 97--121. Cornell
  University Press, Ithaca, NY, 1967.

\end{thebibliography}

\end{document}